\documentclass{article}

\PassOptionsToPackage{numbers, compress}{natbib}

\usepackage[preprint]{neurips_2021}

\usepackage[utf8]{inputenc} 
\usepackage[T1]{fontenc}    
\usepackage{hyperref}       
\usepackage{url}            
\usepackage{booktabs,colortbl}       
\usepackage{amsfonts}       
\usepackage{nicefrac}       
\usepackage{microtype}      

\usepackage{dsfont}
\usepackage{times}
\usepackage{latexsym}
\usepackage{color}
\usepackage{multirow}
\usepackage{forest}
\usepackage{tikz}
\usepackage{algorithm}
\usepackage{algpseudocode}
\usepackage{amsmath}
\usepackage{graphics}
\usepackage{graphicx}
\usepackage{subfig}
\usepackage{mwe}
\usepackage{amssymb}
\usepackage{bm}
\usepackage{amsmath}
\usepackage{makecell}
\usepackage{tabularx}
\usepackage{arydshln}
\usepackage{url}
\usepackage{CJKutf8}
\usepackage{wrapfig}
\usepackage{enumitem}

\definecolor{myLinkColor}{rgb}{0.18,0.39,0.62}
\definecolor{darkColor}{rgb}{0, 0, 0}

\definecolor{rred}{HTML}{c4260b}
\definecolor{bblue}{HTML}{4F81BD}

\hypersetup{
    colorlinks=true,
    linkcolor=myLinkColor,
    filecolor=myLinkColor,
    urlcolor=myLinkColor,
    citecolor=myLinkColor,
}

\definecolor{zred}{RGB}{196, 38, 11}
\definecolor{zblue}{RGB}{41, 52, 190}
\definecolor{zgreen}{RGB}{18, 141, 21}

\definecolor{zptu}{RGB}{18, 141, 21}

\definecolor{shuo}{RGB}{237, 141, 0}

\definecolor{wxwang}{RGB}{18, 21, 141}

\makeatletter
\def\adl@drawiv#1#2#3{%
        \hskip.5\tabcolsep
        \xleaders#3{#2.5\@tempdimb #1{1}#2.5\@tempdimb}%
                #2\z@ plus1fil minus1fil\relax
        \hskip.5\tabcolsep}
\newcommand{\cdashlinelr}[1]{%
  \noalign{\vskip\aboverulesep
           \global\let\@dashdrawstore\adl@draw
           \global\let\adl@draw\adl@drawiv}
  \cdashline{#1}
  \noalign{\global\let\adl@draw\@dashdrawstore
           \vskip\belowrulesep}}
\makeatother

\title{Language Models are Good Translators}

\author{%
Shuo Wang$^1$ Zhaopeng Tu$^2$ Zhixing Tan$^1$ Wenxuan Wang$^2$ Maosong Sun$^{1,3}$ Yang Liu$^{1,3,4}$ \\
$^1$Dept. of Comp. Sci. \& Tech., Institute for AI, BNRist Center, Tsinghua University\\
$^3$Beijing Academy of Artificial Intelligence \ $^4$Institute for AIR, Tsinghua University\\
$^1${\texttt{wangshuo.thu@gamil.com}} \\
$^1${\texttt{\{zxtan, sms, liuyang2011\}@tsinghua.edu.cn}}\\
$^2$Tencent AI Lab\\
$^2${\texttt{\{zptu, jwxwang\}@tencent.com}}\\
}

\begin{document}

\maketitle

\begin{abstract}

Recent years have witnessed the rapid advance in neural machine translation (NMT), 
the core of which lies in the encoder-decoder architecture. 
Inspired by the recent progress of large-scale pre-trained language models on machine translation in a limited scenario, we firstly demonstrate that a single language model (\textsc{Lm4Mt}) can achieve comparable performance with strong encoder-decoder NMT models on standard machine translation benchmarks, using the same training data and similar amount of model parameters.
\textsc{Lm4Mt} can also easily utilize source-side texts as additional supervision.
Though modeling the source- and target-language texts with the same mechanism,
\textsc{Lm4Mt} can provide unified representations for both source and target sentences, which can better transfer knowledge across languages.
Extensive experiments on pivot-based and zero-shot translation tasks show that \textsc{Lm4Mt} can outperform the encoder-decoder NMT model by a large margin.
\end{abstract}

\section{Introduction}
Recent years have witnessed the success of the neural machine translation (NMT) models~\cite{Sutskever:2014:NIPS,Bahdanau:2015:RNNSearch,Gehring:2017:ConvSeq2Seq,Vaswani:2017:Attention}, which translate texts in the source language into the target language with neural networks. 
A number of studies have directed their attention to designing more advanced NMT models. \textsc{Transformer}~\cite{Vaswani:2017:Attention} is the most widely-used NMT model, which uses attention-based neural networks for both the encoder and the decoder. \cite{Wu:2019:LightConv} propose an efficient NMT model using lightweight and dynamic convolutions. \cite{So:2019:Evolved} apply neural architecture search and find a better alternative to vanilla \textsc{Transformer} model.
Although the core mechanism has evolved from RNN to self-attention and then other alternatives, the encoder-decoder architecture is still the dominating framework for NMT models.

Previous studies have shown that the boundary between encoder and decoder, in terms of the localization of the representation in the continuous space, is blurry for multilingual NMT~\cite{kudugunta-etal-2019-investigating}.
~\cite{He:2018:Layer} demonstrate that the standard NMT model benefits from weakening the boundary between encoder and decoder by sharing parameters of the two components.
More recently, GPT-3~\cite{Brown:2020:GPT3}, which is a single language model (LM) pre-trained on huge amount of multilingual data, firstly shows some promising results of LM on machine translation when given in-context translation examples as prefixes. 
GPT-3 has several key limitations that prevent it from serving as the practical NMT model, including (1) GPT-3 fails for machine translation without in-context prefixes; (2) both the amount of training data and model parameters for GPT-3 are several orders of magnitude larger than those of standard NMT models, which is prohibitively expensive for many researchers and developers.
Nevertheless, the surprising results still trigger us to think a research question: {\em can we really accomplish the machine translation task with a single language model?}

To answer this question, we explore the ability of language models for machine translation (i.e., \textproc{Lm4Mt}) with only limited parallel data, which is often used to train the standard encoder-decoder NMT models.
Surprisingly, we find through experiments that the vanilla language model only marginally underperforms the encoder-decoder NMT models with comparable model sizes. 
Benefiting from the characteristic of \textsc{Lm4Mt} to generate both the source and target sentences in the same manner, we introduce an auto-encoding loss with a decaying schedule to help \textproc{Lm4Mt} better learn from source-language sentences. 
Experimental results show that the proposed \textproc{Lm4Mt} achieves comparable performance with or even better performance than its encoder-decoder counterpart on six machine translation benchmarks.
For instance, on the benchmarking WMT14 English$\Rightarrow$German and English$\Rightarrow$French translation tasks, \textproc{Lm4Mt} achieves 29.3 BLEU and 42.9 BLEU, respectively.
The additional source-side supervision can improve the model performance in two other aspects: (1) higher translation quality for source-original sentences, which are usually more complex and difficult to translate~\cite{Zhang:2019:Translationese}; and (2) better model robustness against missing word perturbations. These results reveal that the source-side supervision can help \textsc{Lm4Mt} better understand source texts.

Another appealing advantage of \textsc{Lm4Mt} is the unified representation for both source and target sentences, which might better transfer knowledge across languages. We empirically validate our hypotheses in two scenarios: (1) {\em pivot-based translation} where a pivot language serves as the transit station to transfer the knowledge from the source language to the target language~\cite{Kim:2019:Pivot}; and (2) {\em zero-shot translation}\footnote{Zero-shot translation denotes translating between language pairs that do not exist in the training data.} for multilingual NMT model, which has a stricter requirement on the model representations to implicitly bridge between the language pairs unseen during training~\cite{johnson-etal-2017-googles}. Experimental results show that \textsc{Lm4Mt} can outperform the encoder-decoder NMT model in all cases.
We find the source-side auto-encoding loss is essential for \textsc{Lm4Mt} to perform zero-shot translation.

To sum up, the main contributions of this paper are listed as follows:
\begin{itemize}
    \item We firstly demonstrate that a single language model can achieve comparable translation performance with the encoder-decoder NMT model of the same model size.
    
    \item Benefiting from the additional source-side supervision and unified representations across different languages, the proposed \textsc{Lm4Mt} can outperform the encoder-decoder NMT model in both pivot-based and zero-shot translation scenarios.
\end{itemize}

\section{Related Work}

\paragraph{Functionalities and Importance of Encoder}
In recent years, there has been a growing interest in understanding the functionalities of the encoder in encoder-decoder NMT models.
For example, \cite{Tang:2019:Enc-Free} simplify the \textsc{Transformer} model to an encoder-free model and find that the encoder is crucial for NMT models to achieve good results.
\cite{Wang:2019:Deep,Wang:2020:Calibration} show that enlarging the capacity of the encoder is more effective for improving translation performance than enlarging the decoder. However, these conclusions might only hold for the encoder-decoder architecture. In this paper, we rethink the importance of encoder in a new architecture and reveal that a single decoder can accomplish the translation task well.

\paragraph{Shared Encoder and Decoder}
\cite{kudugunta-etal-2019-investigating} find that the boundary between encoder and decoder is blurry for multilingual NMT.
In standard translation, there are some works that weaken the boundary between the encoder and decoder. For example,~\cite{He:2018:Layer} share the parameters of the encoder and decoder, which coordinates the learning of hidden representations of the two components. 
However, their model still encodes source-language texts in the same way as vanilla encoder-decoder NMT models, ignoring the source-side supervision. The source and target sentences are still consumed in separate mechanisms. We take one step further and simplify the architecture into a simple decoder, where we can easily utilize the source-side supervision to help better understand source texts.

\paragraph{Language Model for Machine Translation}
Recently, the pre-trained language model GPT-3~\cite{Brown:2020:GPT3} has shown encouraging results on machine translation tasks. GPT-3 is pre-trained on massive data with a huge amount of parameters, and requires in-context translation examples to achieve good translation performance. 
Our work, on the other hand, is the first attempt to systematically compare the ability of encoder-decoder models and LM (i.e., single decoder) across different translation scenarios, using the same training data and similar model sizes.

\section{Approach}

\subsection{Preliminaries}

\paragraph{Language Model}
Given a monolingual sentence $\bm{\mathrm{y}}=\left \{ y_1, y_2, \dots, y_T \right \}$, the goal of language modeling is to estimate the joint probability $P(\bm{\mathrm{y}})$, which is usually auto-regressively factorized as
\begin{equation}
    \label{eq:lm}
    P(\bm{\mathrm{y}}) = \prod_{t=1}^{T} P(y_t | \bm{\mathrm{y}}_{<t}),
\end{equation}
where $\bm{\mathrm{y}}_{<t} = \{y_1, y_2, \dots, y_{t-1}\}$ is the prefix before $y_t$.
With this factorization, the problem is simplified to estimating each conditional factor. Standard neural language models~\cite{Dai:2019:TransformerXL} encode the context $\bm{\mathrm{y}}_{<t}$ into a continuous vector, which is then multiplied by the word embedding matrix to obtain the logits. The logits are then used to compute the probability distribution over the next word through the Softmax function.

\paragraph{Encoder-Decoder NMT Model}
Given a source-language sentence $\bm{\mathrm{x}}=\left \{ x_1, x_2, \dots, x_S \right \}$, NMT models learn to predict the conditional probability of the corresponding target-language sentence $\bm{\mathrm{y}}$:
\begin{equation}
    \label{eq:mt}
    P(\bm{\mathbf{y}} | \bm{\mathbf{x}}) = \prod_{t=1}^{T} P(y_t | \bm{\mathrm{x}}, \bm{\mathrm{y}}_{<t}).
\end{equation}

Most previous NMT models use the encoder-decoder framework~\cite{Bahdanau:2015:RNNSearch,Vaswani:2017:Attention,Liu:2020:mBART}. The encoder is a feature extractor, which maps the source-language sentence $\bm{\mathrm{x}}$ into a sequence of continuous representations $\bm{\mathrm{r}} = \left\{\bm{\mathrm{r}}_1, \bm{\mathrm{r}}_2, \dots, \bm{\mathrm{r}}_{S}\right\}$. The decoder is a conditional language model, which estimates the probability $P(\bm{\mathrm{y}} | \bm{\mathrm{r}})$. In the widely used \textsc{Transformer}~\cite{Vaswani:2017:Attention} model, both the encoder and the decoder use attention networks, which are shown effective to learn contextualized representations~\cite{Devlin:2019:BERT}. To achieve good translation performance, NMT models are expected to not only extract effective source-language representations $\bm{\mathrm{r}}$, but also be capable of generating a fluent target-language sentence that can recover the information conveyed in the source sentence.

\begin{figure*}[t]
\centering
\subfloat[encoder-decoder NMT model]{\includegraphics[height=0.228\textwidth]{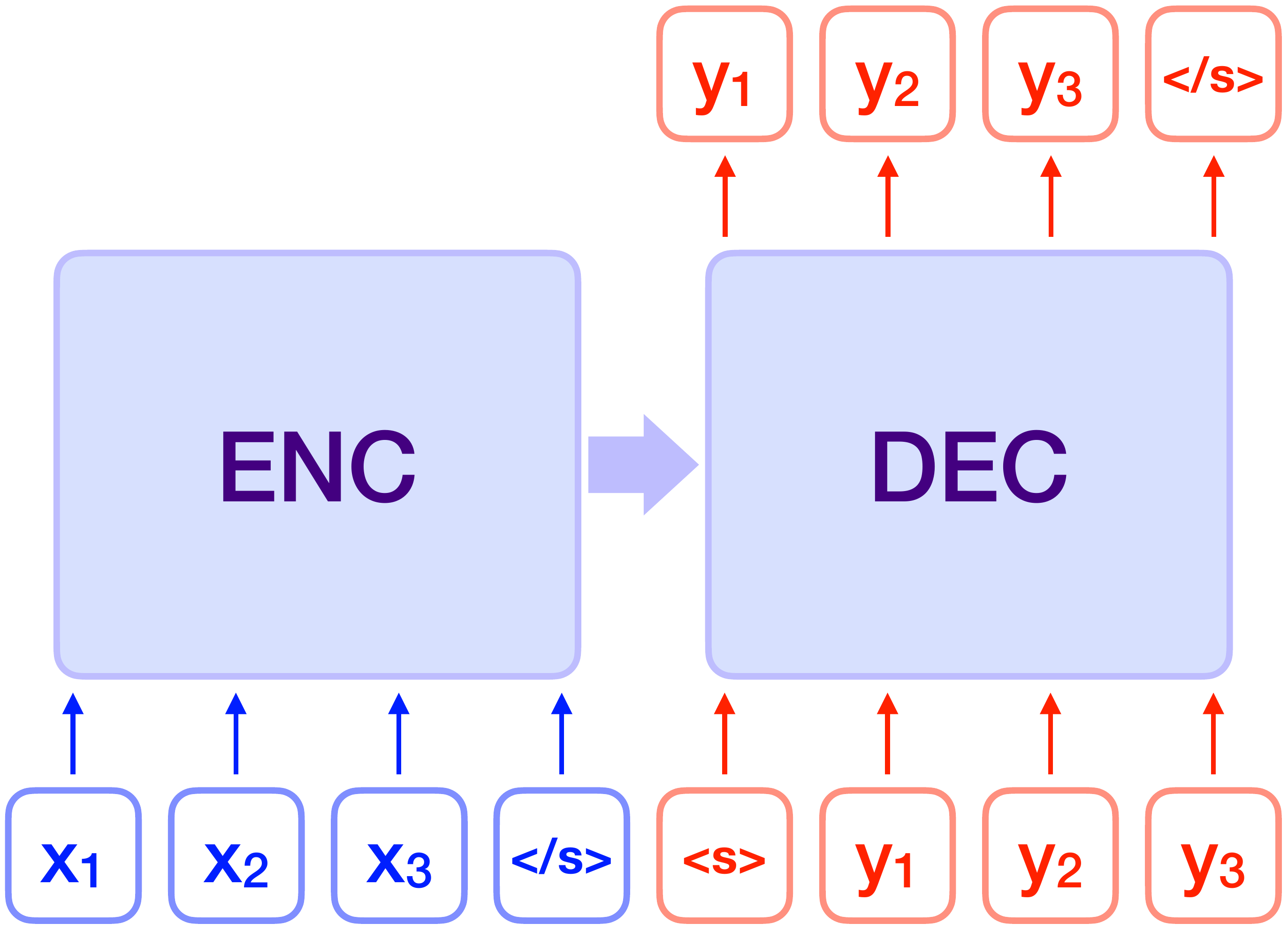}}
\hspace{0.02\textwidth}
\subfloat[\textproc{Lm4Mt} during training]{\includegraphics[height=0.228\textwidth]{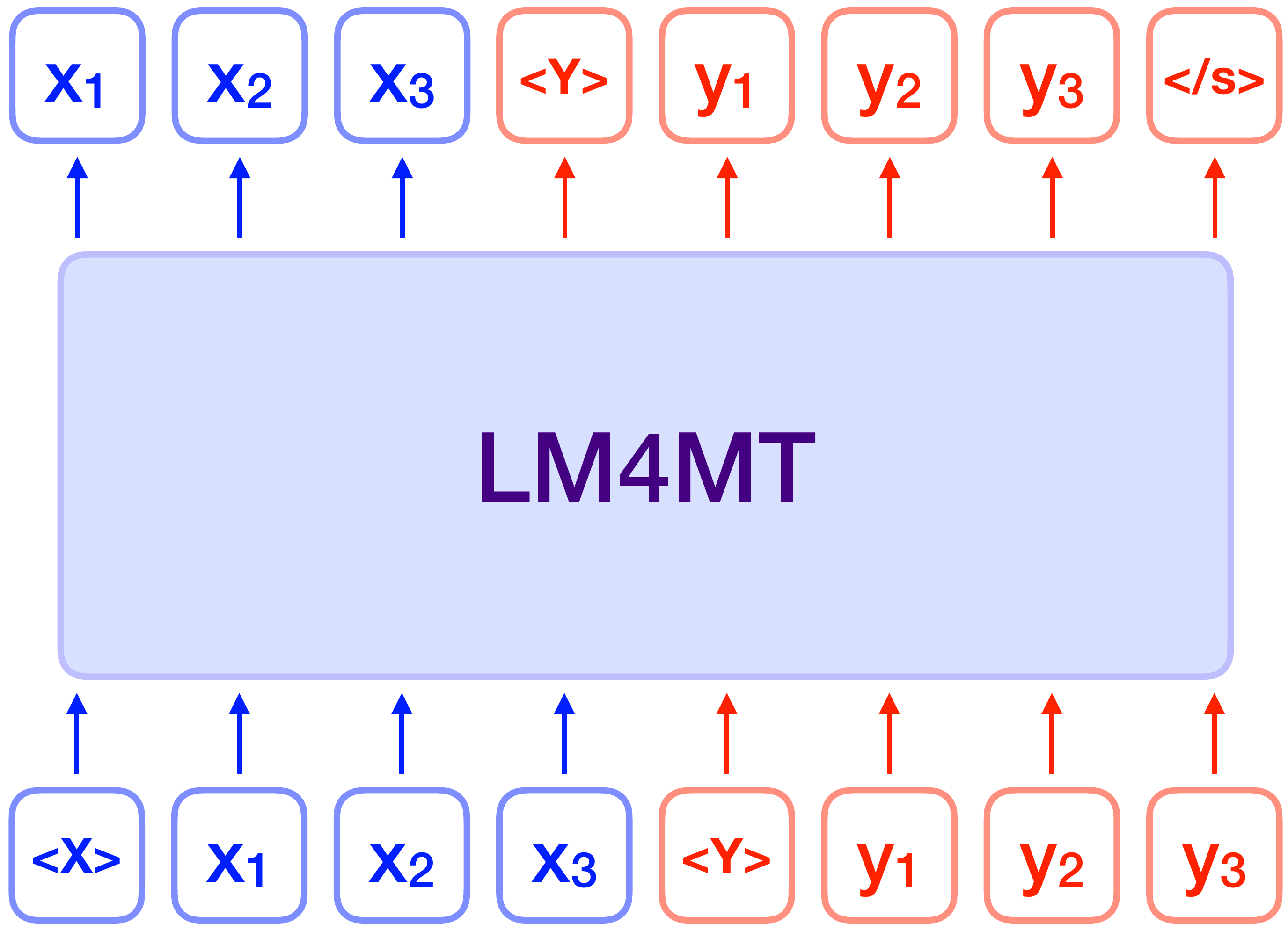}\label{fig:lm4mt-train}}
\hspace{0.02\textwidth}
\subfloat[\textproc{Lm4Mt} during inference]{\includegraphics[height=0.228\textwidth]{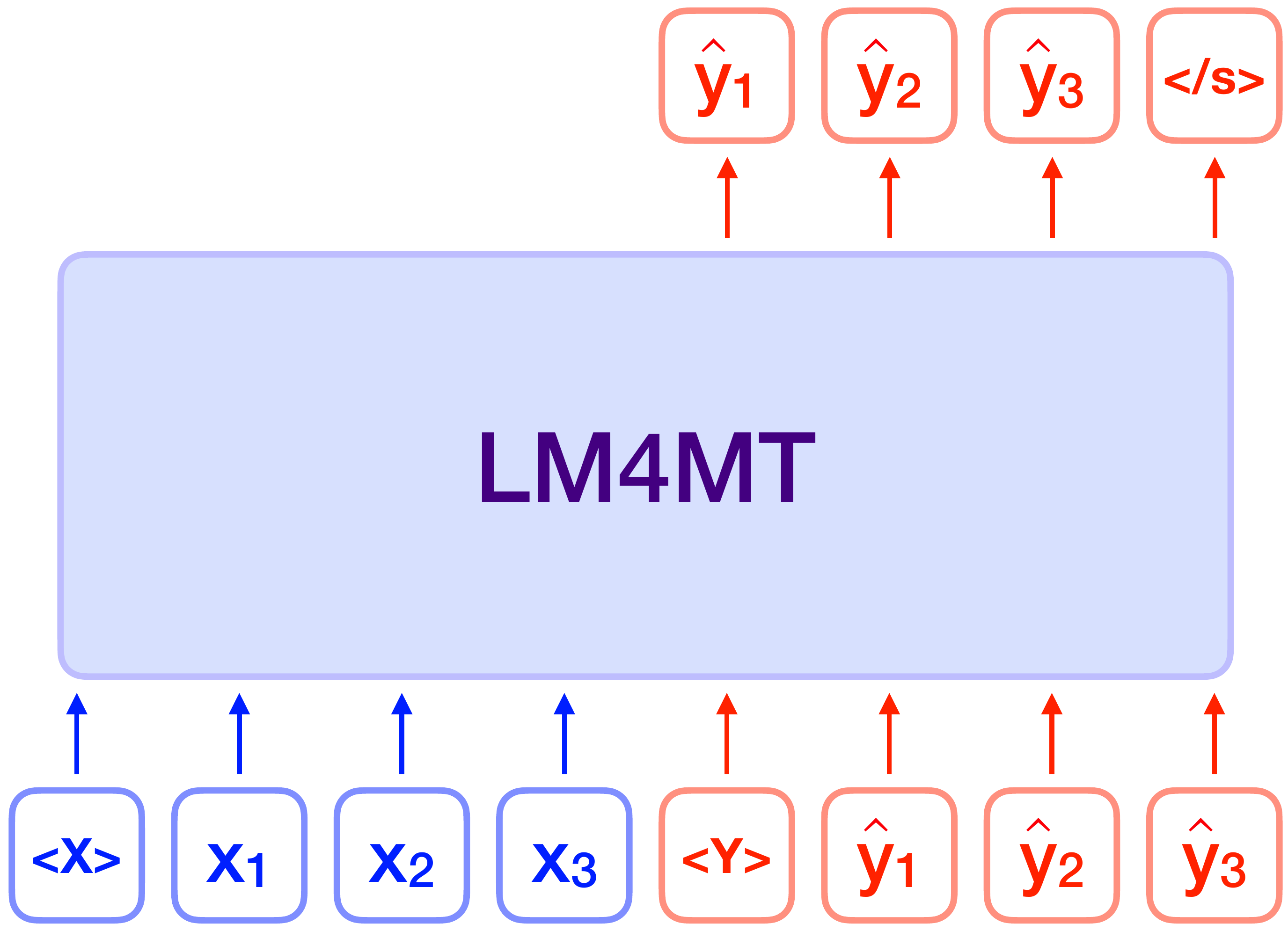}\label{fig:lm4mt-inf}}
\caption{Illustration of \textproc{Lm4Mt}. For comparison, we also plot the encoder-decoder model. During training, we feed \textproc{Lm4Mt} with the concatenation of the source- and target-language sentences, which are explicitly separated by special language tags. At the inference time, the source-language text, together with the target-language tag, is used as the prefix for \textproc{Lm4Mt} to generate the translation.}
\end{figure*}

\subsection{Language Model for Machine Translation}
\label{subsec: model}
Recently, several works find that huge language models pre-trained on large-scale data set can achieve good results on a number of natural language understanding and generation tasks~\cite{Radford:2019:GPT-2,Brown:2020:GPT3,Du:2021:GLM,liu:2021:GPT-U}, indicating the potential of language models to serve as a good feature extractor. However, for machine translation, it is still not well investigated whether a language model along can act as a source-language feature extractor and a target-language generator at the same time, especially without huge model size and large amount of training data. In this work, we aim to investigate the capability of language models to perform machine translation, which is a task that requires the abilities of both language understanding and generation. 

From Equation~(\ref{eq:lm})~and~(\ref{eq:mt}), we find that objectives of language modeling and machine translation are quite similar, since the source-language sentence $\bm{\mathrm{x}}$ can be seen as a special type of prefix. Inspired by this observation, we propose the language model for machine translation (i.e., \textproc{Lm4Mt}) that are trained to estimate the joint probability of the two sentence $\bm{\mathrm{x}}$ and $\bm{\mathrm{y}}$:
\begin{equation}
    P(\bm{\mathrm{x}}, \bm{\mathrm{y}}) = \prod_{s=1}^{S} P(x_s | \bm{\mathrm{x}}_{<s})
    \prod_{t=1}^{T} P(y_t | \bm{\mathrm{x}}, \bm{\mathrm{y}}_{<t}).
\end{equation}

Specifically, we concatenate $\bm{\mathrm{x}}$ and $\bm{\mathrm{y}}$ into a sentence pair, and then use \textproc{Lm4Mt} to estimate the joint probability of such a sentence pair as if it is one sentence. To help the model better identify the boundary of sentences from different languages, we add a special language tag before each sentence. Figure~\ref{fig:lm4mt-train} depicts an example for the training phrase of \textproc{Lm4Mt}. Experiments in Section~\ref{sec:ablation} show that the added language tags are effective to improve the translation performance of \textproc{Lm4Mt}.

\paragraph{Training} Just as standard language models, the training objective of \textproc{Lm4Mt} is to minimize the negative log-likelihood of the probability $P(\bm{\mathrm{x}}, \bm{\mathrm{y}})$:
\begin{equation}
\begin{split}
    -\log P(\bm{\mathrm{x}}, \bm{\mathrm{y}}) = \mathcal{L}^{\mathrm{\textproc{AE}}} + \mathcal{L}^{\mathrm{\textproc{MT}}} = - \sum_{s=1}^{S} \log P(x_s | \bm{\mathrm{x}}_{<s}) - \sum_{t=1}^{T} P(y_t | \bm{\mathrm{x}}, \bm{\mathrm{y}}_{<t}),
\end{split}
\end{equation}
where $\mathcal{L}^{\mathrm{\textproc{AE}}}=- \sum_{s=1}^{S} \log P(x_s | \bm{\mathrm{x}}_{<s})$, which is the auto-encoding loss, reflecting the ability of the model to reconstruct the source-language sentence $\bm{\mathrm{x}}$. $\mathcal{L}^{\mathrm{\textproc{MT}}} = - \sum_{t=1}^{T} P(y_t | \bm{\mathrm{x}}, \bm{\mathrm{y}}_{<t})$, which is the machine translation loss that has been widely-used for NMT models. Compared to encoder-decoder NMT models, \textproc{Lm4Mt} is trained with an additional source-side auto-encoding loss $\mathcal{L}^{\mathrm{\textproc{AE}}}$.

Intuitively, the explicit supervision induced by $\mathcal{L}^{\mathrm{\textproc{AE}}}$ may help \textproc{Lm4Mt} better understand the source-language sentences. Moreover, using $\mathcal{L}^{\mathrm{\textproc{AE}}}$ makes the modeling mechanisms of the source and target texts more similar, which may reduce the representation gap between source and target sentences. We find through experiments (Section~\ref{sec:ablation}) that simply adding $\mathcal{L}^{\mathrm{\textproc{AE}}}$ is not a good practice for machine translation, which might be caused by that $\mathcal{L}^{\mathrm{\textproc{AE}}}$ prevents the model to further minimize $\mathcal{L}^{\mathrm{\textproc{MT}}}$ at the end of the training. Inspired by the training strategy of unsupervised NMT~\cite{Lample:2018:UNMT,Lample:2019:XLM}, we multiply $\mathcal{L}^{\mathrm{\textproc{AE}}}$ with a decaying factor $\lambda_d$. Therefore, the training objective of \textproc{Lm4Mt} is
\begin{equation}
    \label{eq:l-lm4mt}
    \mathcal{L}^{\mathrm{\textproc{Lm4Mt}}} = \lambda_d \mathcal{L}^{\mathrm{\textproc{AE}}} + \mathcal{L}^{\mathrm{\textproc{MT}}}.
\end{equation}

\begin{wrapfigure}{r}{6cm}
\vspace{-1pt}
\centering
\includegraphics[width=0.36\textwidth]{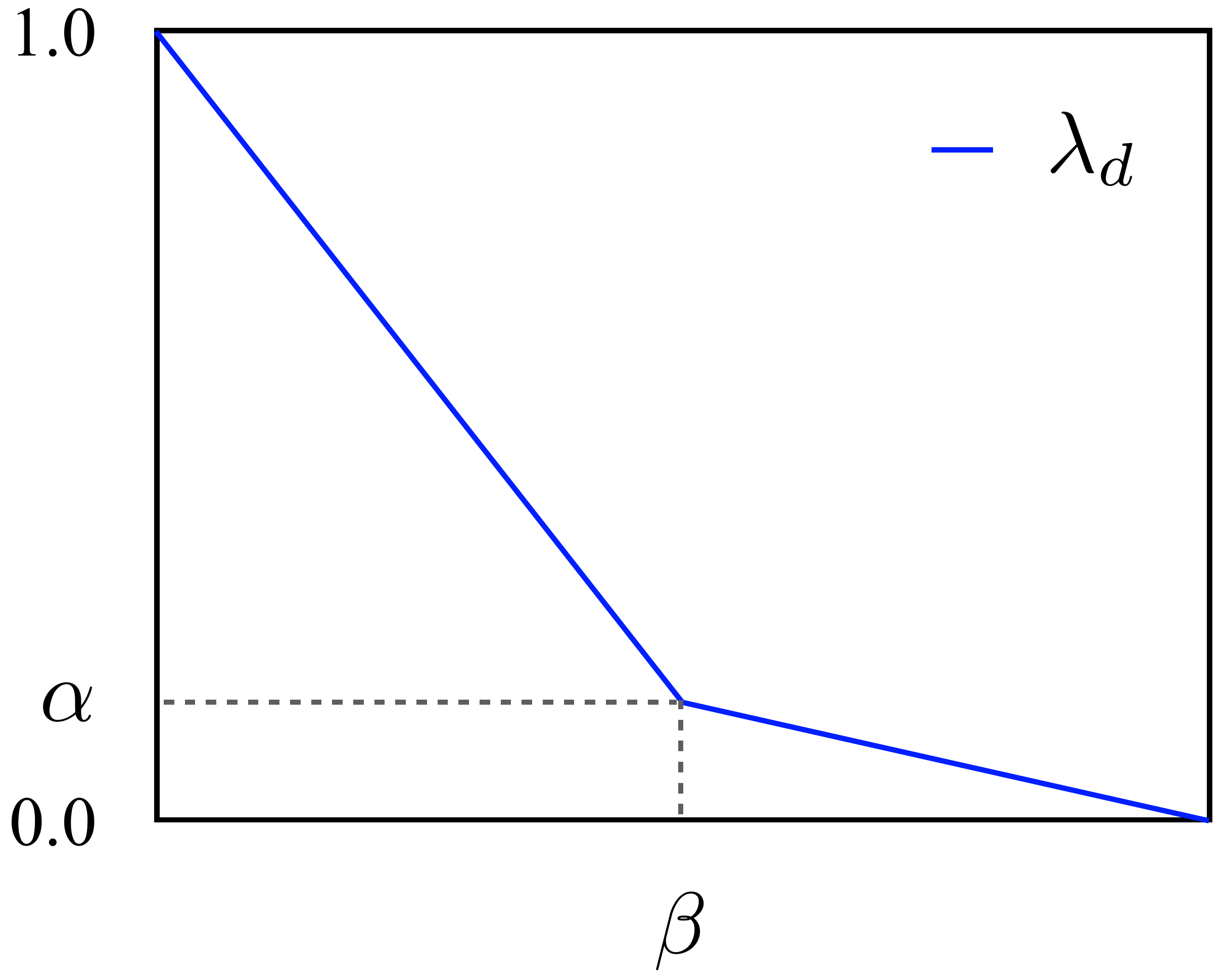}
\caption{Decaying schedule for $\lambda_d$ in Eq.~(\ref{eq:l-lm4mt}). $\alpha$ and $\beta$ are hyper-parameters.}
\label{fig:lambda}
\vspace{-25pt}
\end{wrapfigure}
As shown in Figure~\ref{fig:lambda}, we use a piece-wise linear decaying schedule for the factor $\lambda_d$. Through this strategy, we expect \textproc{Lm4Mt} to better learn from the source sentences at early stages and then focus more on the translation loss at the end of the training.

\paragraph{Inference}
Different from the training phrase during which \textproc{Lm4Mt} estimates the probability for both the source- and target-language sentences, we only let \textproc{Lm4Mt} predict the target-language tokens at the inference time. As shown in Figure~\ref{fig:lm4mt-inf}, \textproc{Lm4Mt} takes in source-language tokens and encodes them into hidden states in parallel. Then the proposed model generates each target-language token step by step.

\section{Experimental Setup}

\paragraph{Data}
To make a thorough comparison between the widely-used \textsc{Transformer}~\cite{Vaswani:2017:Attention} model and our proposed \textproc{Lm4Mt}, we conducted experiments on datasets with different sizes: WMT16\footnote{\url{http://statmt.org/wmt16/translation-task.html}} English-Romanian (En-Ro), WMT14\footnote{\url{http://statmt.org/wmt14/translation-task.html}} English-German (En-De) and English-French (En-Fr), which consist of 0.6M, 4.5M and 35.8M sentence pairs respectively. In En-Ro, we used news-dev2016 as the validation set and news-test2016 as the test set. In En-De and En-Fr, we used news-test2013 and news-test2014 as the validation and test sets, respectively.
We applied BPE~\cite{Sennrich:2016:bpe} with 32K merge operations for En-De data, and with 40K merge operations for En-Ro and En-Fr data.

\paragraph{Model}
We used the encoder-decoder based \textsc{Transformer} model~\cite{Vaswani:2017:Attention} as our baseline. We used models of two sizes, namely the \textsc{Transformer}-Base and \textsc{Transformer}-Big, both of which consist of a 6-layer encoder and a 6-layer decoder. The hidden sizes of \textsc{Transformer}-Base and \textsc{Transformer}-Big are 512 and 1024, respectively. We also list the results of some recent representative architectures for comparison, including \textsc{Transformer} with relative position representations~\cite{Shaw:2018:RelPos}, scaling \textsc{Transformer}~\cite{Ott:2018:Scale}, layer-wise coordinated \textsc{Transformer}~\cite{He:2018:Layer}, dynamic convolutions~\cite{Wu:2019:LightConv}, and evolved \textsc{Transformer}~\cite{So:2019:Evolved}.

The proposed \textproc{Lm4Mt} model consists of only a self-attention decoder. There are mainly three differences between the \textproc{Lm4Mt} model and the decoder of the vanilla \textsc{Transformer}~\cite{Vaswani:2017:Attention}. Firstly, each \textproc{Lm4Mt} layer has only one type of attention network while the decoder in \cite{Vaswani:2017:Attention} has both self-attention and cross-attention networks. Secondly, we use pre-norm residual unit~\cite{Wang:2019:Deep} in order to train deep language models. Thirdly, we follow \cite{Radford:2019:GPT-2} to use GELU activations~\cite{Hendrycks:2016:GELU} in \textproc{Lm4Mt}, which has been shown to be effective in language model training. Similar to \textsc{Transformer}, we also conduct experiments with base and big settings for \textproc{Lm4Mt}. The hidden size of \textproc{Lm4Mt}-Base is 512 and that of \textproc{Lm4Mt}-Big is set to 1024.

\paragraph{Training Details}
For En-Ro, we only trained base models since the training corpus was too small (0.6M sentence pairs) to learn big models. The dropout rate was set to 0.3 for both \textsc{Transformer} and \textproc{Lm4Mt}. We set weight decay to $1$e$-4$ to overcome over-fitting. For En-De, the dropout rates were set to 0.1 for base models and 0.3 for big models. For En-Fr, the dropout rate was set to 0.1 for both base and big models. For all the pivot-based and multilingual translation models, the dropout rate was set to 0.1.
For all the three language pairs, we used Adam~\cite{Kingma:2015:Adam} to optimize the model parameters, with $\beta_1 = 0.9$, $\beta_2 = 0.98$, and $\epsilon = 10^{-9}$. We trained base models on mini-batches that contain approximately 32K target-language tokens for 150K steps. For big models, we followed \cite{Ott:2018:Scale} to use {\em larger batches}, which contain approximately 460K tokens, to further boost the performance of big models. When using the large batch size, we followed \cite{Wu:2019:LightConv} to use the cosine learning rate schedule, where the learning rate was warmed up linearly to $1$e$-3$ in the first 10K steps and then decayed to $1$e$-7$ following a cosine rate within a single cycle. Big models were trained for 30K steps on large batches. We obtained the final model by averaging the last 5 checkpoints, which were saved at 1000-update and 500-update intervals for base and big models, respectively. As for the decaying schedule, we set $\alpha$=0.1, $\beta$=37.5K for base models and $\alpha$=0.1, $\beta$=22.5K for big models.
All models were implemented on the top of {\tt fairseq} toolkit.\footnote{\url{https://github.com/pytorch/fairseq}}
We conducted all the experiments on 8 Nvidia Telsa V100 32GB GPUs.

\section{Experimental Results}

\subsection{Ablation Study}
\label{sec:ablation}


\begin{wraptable}{r}{5.9cm}
    \vspace{-12pt}
    \caption{Effect of language tag (``Tag'') for 19-layer \textsc{Lm4Mt} with $\mathcal{L}^{\mathrm{\textproc{MT}}}$ loss.}
    \label{tab:ablation-lang-tag}
    \centering
    \begin{tabular}{cc c}
    \toprule
    \bf Model & \bf Tag & \bf BLEU \\
    \midrule
    {\textproc{Transformer}} & - & \bf 26.7 \\
    \midrule
    \multirow{2}{*}{\textproc{Lm4Mt}} & \texttimes & 25.8 \\
    & \checkmark & \em 26.3  \\
    \bottomrule
    \end{tabular}
\end{wraptable}

\paragraph{Effect of Language Tag} In the training phase, we simply feed \textproc{Lm4Mt} with the concatenation of the source- and target-language sentences, which may make it difficult to identify the start position of the target-language sentence. In response to this problem, we use language tags to explicitly distinguish sentences of different languages, as shown in Figure~\ref{fig:lm4mt-train}. Table~\ref{tab:ablation-lang-tag} shows the impact of language tags, indicating the importance of explicit indicators of languages for \textproc{Lm4Mt}. In the following experiments we use language tags by default.

\paragraph{Effect of Layer Number}
Since \textsc{Transformer} has an additional encoder compared to \textproc{Lm4Mt}, we deepen the \textproc{Lm4Mt} to assimilate the parameter count. Table~\ref{tab:ablation-layer} shows the results of \textproc{Lm4Mt} with different depths. The hidden size of all models are 512.
Surprisingly, the 6-layer \textproc{Lm4Mt} performs only 1.8 BLEU lower than the encoder-decoder baseline, although it has much fewer parameters (56.9M vs. 98.8M). Enlarging the \textproc{Lm4Mt} by adding layers is effective to improve the translation performance.
When using comparable amount of parameters, the performance of \textproc{Lm4Mt} is 0.4 BLEU point lower than the \textsc{Transformer} baseline (26.3 vs. 26.7 for L19 \textsc{Lm4Mt}). 
In the following experiments, we use L19 as the default architecture for \textproc{Lm4Mt}.

\begin{wraptable}{r}{7cm}
    \vspace{-12pt}
    \caption{Effect of layer number. All models use $\mathcal{L}^{\mathrm{\textproc{MT}}}$ as the training loss.}
    \label{tab:ablation-layer}
    \centering
    \begin{tabular}{ccr c}
    \toprule
    \bf Model & \bf Arch. & \bf Param. & \bf BLEU \\
    \midrule
    {\textproc{Transformer}} & {L6-L6} & {98.8M} & \bf 26.7 \\
    \midrule
    \multirow{4}{*}{\textproc{Lm4Mt}} & L6~~ & 56.9M & 24.9 \\
    & L12 & 75.8M & 25.6 \\
    & L18 & 94.7M & 26.2 \\
    & L19 & 97.8M & \em 26.3 \\
    \bottomrule
    \end{tabular}
\end{wraptable}

\paragraph{Effect of Auto-Encoding Loss}
Another key difference between \textproc{Lm4Mt} and \textsc{Transformer} is that we introduce an additional auto-encoding loss $\mathcal{L}^{\mathrm{\textproc{AE}}}$ (Equation~(\ref{eq:l-lm4mt})) for \textproc{Lm4Mt} to better understand the source sentence. To empirically validate the effect of the source-side $\mathcal{L}^{\mathrm{\textproc{AE}}}$, we follow~\cite{Graham:2019:hallu} to separately report the BLEU score on the source-original\footnote{Source-original texts are written by source-language native speakers, which are found to be more difficult to translate than human-translated texts~\cite{Zhang:2019:Translationese}. Since the En$\Rightarrow$De validation set contains six different original languages, we report the source-original results on the En$\Rightarrow$De test set.} sentences on WMT14 En$\Rightarrow$De test set. Intuitively, the translation of source-original sentences requires a better understanding of source sentences to handle the relatively more complex sentence structures and more diverse contents~\cite{Wang:2021:LCB}.

As listed in Table~\ref{tab:ablation-ae-loss}, directly adding auto-encoding loss (``$\mathcal{L}^{\mathrm{\textproc{AE}}}+\mathcal{L}^{\mathrm{\textproc{MT}}}$'') fails to improve translation performance. We plot the learning curves of validation perplexities for \textproc{Lm4Mt} trained with different objectives. We find that directly adding the auto-encoding loss inversely increase the validation perplexity in late training stages.
Using the decaying schedule to adjust the weight of $\mathcal{L}^{\mathrm{\textproc{AE}}}$ can effectively improve translation performance and help training convergence.
The performance improvement is mainly from better translation of source-original sentences (``o''), which confirms our claim that the auto-encoding loss can help to better understand the source sentences.
In the following experiments, we train \textproc{Lm4Mt} models with decaying auto-encoding loss by default.

\vspace{5pt}

\begin{minipage}{\textwidth}
  \begin{minipage}[t]{0.56\textwidth}
    \centering
    \captionof{table}{Effect of auto-encoding loss $\mathcal{L}^{\mathrm{\textproc{AE}}}$. 
    $\lambda_d$ denotes a dynamic weight decaying from 1 to 0 during training. ``o'' denotes source-original sentences in the test set, while ``n-o'' denotes the target-original sentences.}
    \setlength{\tabcolsep}{3.5pt}
    \label{tab:ablation-ae-loss}
    \begin{tabular}{c r cccc}
    \toprule
    \multirow{2}{*}{\bf Model} & \multirow{2}{*}{\bf  Loss}  &   \multirow{2}{*}{\bf Valid} & \multicolumn{3}{c}{\bf Test} \\
    \cmidrule(lr){4-6}
            &   &   &  \bf all  &   \bf o    &   \bf n-o \\
    \cmidrule(lr){1-1} \cmidrule(lr){2-2} \cmidrule(lr){3-3} \cmidrule(lr){4-6}
    {\textproc{Trans.}} & $\mathcal{L}^{\mathrm{\textproc{MT}}}$               & 26.7 & \cellcolor{red!0} \bf 27.8 & \cellcolor{red!0} 27.4 & \cellcolor{red!0} \bf 28.1 \\
    \cmidrule(lr){1-1} \cmidrule(lr){2-2}  \cmidrule(lr){3-3} \cmidrule(lr){4-6}
    \multirow{3}{*}{\textproc{Lm4Mt}} & $\mathcal{L}^{\mathrm{\textproc{MT}}}$ & 26.3 & \cellcolor{red!0} 27.3 & \cellcolor{red!0} 26.5 & \cellcolor{red!0} 27.6 \\
    & $\mathcal{L}^{\mathrm{\textproc{AE}}}+\mathcal{L}^{\mathrm{\textproc{MT}}}$ & 26.0 & \cellcolor{zptu!20} 26.9 & \cellcolor{red!5} 26.6 & \cellcolor{zptu!30} 27.0 \\
    & $\lambda_d\mathcal{L}^{\mathrm{\textproc{AE}}}+\mathcal{L}^{\mathrm{\textproc{MT}}}$       & \bf 26.8 & \cellcolor{red!15} \bf 27.8 &  \cellcolor{red!55} \bf 27.6 &  \cellcolor{red!15} 27.9 \\
    \bottomrule
    \end{tabular}
  \end{minipage}
  \hfill
  \begin{minipage}[t]{0.38\textwidth}
    \centering
    \label{fig:tgt-ppl}
    \includegraphics[width=\textwidth]{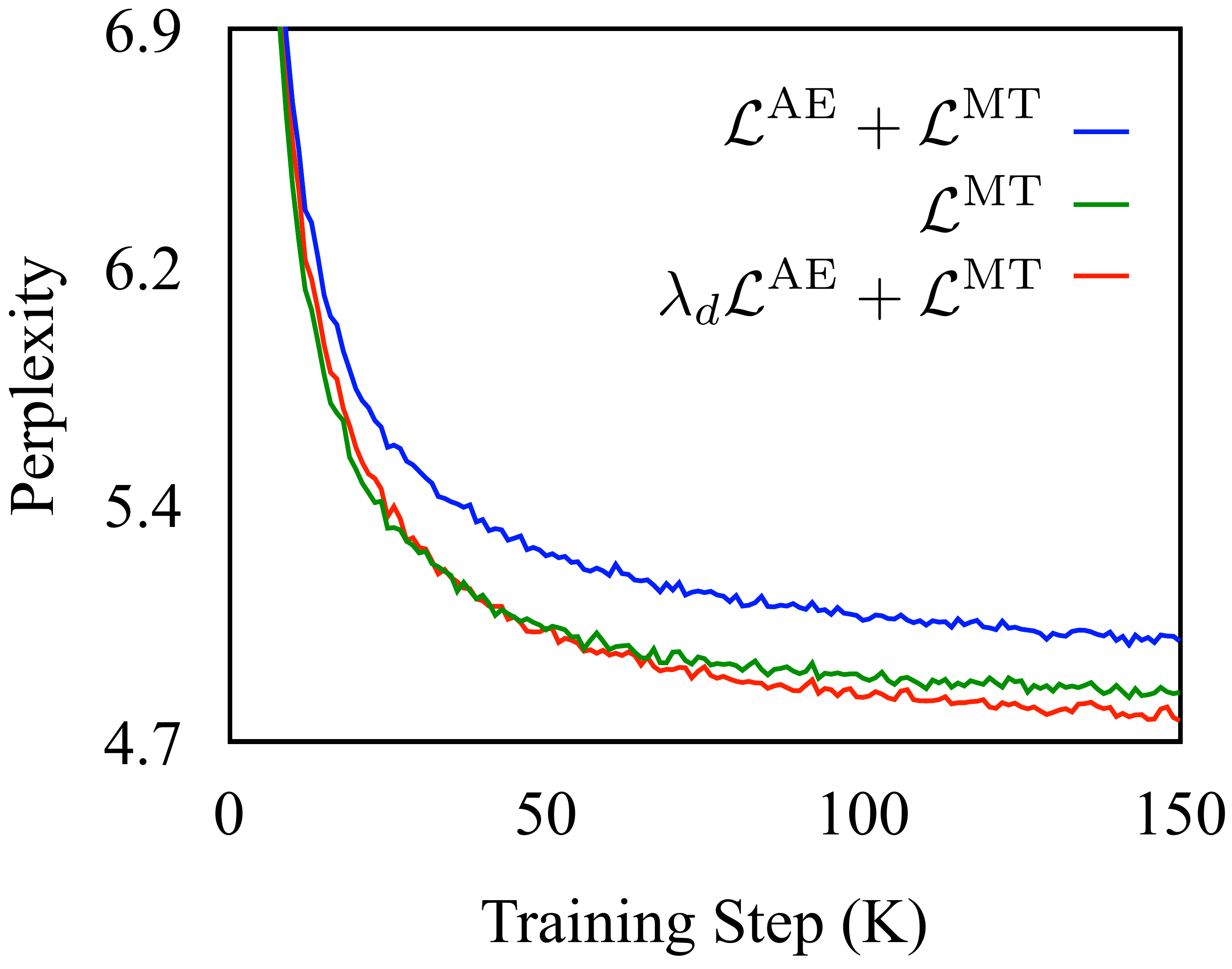}
    \captionof{figure}{Learning curves of validation perplexity on WMT14 En$\Rightarrow$De.}
  \end{minipage}
\end{minipage}

\subsection{Standard Translation}

\begin{table}[t]
    \caption{Translation performance on WMT16 En$\Leftrightarrow$Ro (0.6M sentence pairs), WMT14 En$\Leftrightarrow$De (4.6M sentence pairs) and WMT14 En$\Leftrightarrow$Fr (35.8M sentence pairs) test sets.}
    \label{tab:main-res}
    \centering
    \setlength{\tabcolsep}{3.8pt}
    \begin{tabular}{r cc cc cc}
    \toprule
    \multirow{2}{*}{\bf Model} & \multicolumn{2}{c}{\bf WMT16 En$\Leftrightarrow$Ro} & \multicolumn{2}{c}{\bf WMT14 En$\Leftrightarrow$De} & \multicolumn{2}{c}{\bf WMT14 En$\Leftrightarrow$Fr}\\
    \cmidrule(lr){2-3} \cmidrule(lr){4-5} \cmidrule(lr){6-7}
    & \bf En$\Rightarrow$Ro & \bf Ro$\Rightarrow$En   & \bf En$\Rightarrow$De & \bf De$\Rightarrow$En  &  \bf En$\Rightarrow$Fr  &  \bf Fr$\Rightarrow$En\\
    \midrule
    \multicolumn{7}{c}{\bf Existing Work}\\
    \textproc{Transformer-Base}~\cite{Vaswani:2017:Attention} & - & - & 27.3 & - & 38.1 & - \\
    \textproc{Transformer-Big}~\cite{Vaswani:2017:Attention} & - & - & 28.4 & - & 41.0 & - \\
    \textproc{Relative Position}~\cite{Shaw:2018:RelPos} & - & - & 29.2 & - & 41.5 & - \\
    \textproc{Scale-Trans.}~\cite{Ott:2018:Scale} & - & - & 29.3 & - & 43.2 & - \\
    \textproc{Layer-wise Coor.}~\cite{He:2018:Layer} & 34.4 & - & 29.0 & - & - & - \\
    \textproc{Dynamic Conv.}~\cite{Wu:2019:LightConv} & - & - & 29.7 & - & 43.2 & - \\
    \textproc{Evolved-Trans.}~\cite{So:2019:Evolved} & - & - & 29.5 & - & 41.3 & - \\
    \midrule
    \multicolumn{7}{c}{\bf Our Implementation}\\
    \textproc{Transformer-Base} & \bf 34.4  & 33.9 & \bf 27.8 & 31.3 & \bf 41.2 & 36.5 \\
    \textproc{\textproc{Lm4Mt}-Base}       & 34.2 & \bf 34.1 & \bf 27.8 & \bf 31.5 & \bf 41.2 & \bf 37.1 \\
    \midrule
    \textproc{Transformer-Big} & - & - & \bf 29.3 & 33.0 & \bf 43.0 & 39.0 \\
    \textproc{\textproc{Lm4Mt}-Big} & - & - & \bf 29.3 & \bf 33.2 & 42.9 & \bf 39.5 \\
    \bottomrule
    \end{tabular}
\end{table}

\paragraph{Translation Performance}
Table~\ref{tab:main-res} lists the results on several WMT benchmarks with different data scales. 
\textproc{Lm4Mt} performs very competitively on all the six translation directions, demonstrating that our approach is applicable to low-, medium-, and high-resource language pairs. For instance, on the widely-used WMT14 En$\Rightarrow$De benchmark, \textproc{Lm4Mt-Big} performs as well as our implemented \textsc{Transformer-Big}, which is also comparable with several strong baselines. Note that our work is complementary to many previous works since \textproc{Lm4Mt} can use more advanced components, such as dynamic convolution~\cite{Wu:2019:LightConv} and evolved \textsc{Transformer} cell~\cite{So:2019:Evolved}.
Closely related to our work, \textproc{Layer-wise Coor.}~\cite{He:2018:Layer} shares the parameters of the encoder and decoder, but they still use separate mechanisms to model the source- and target-language texts and ignore the source-side supervision. We take one step further and simplify the architecture into a simple decoder.

\begin{wraptable}{r}{7cm}
    \caption{Translation performance with missing words. All models use the base setting.}
    \label{tab:miss-word}
    \centering
    \begin{tabular}{c ccc}
    \toprule
    \bf Missing Ratio & \bf 0\% & \bf 30\% & \bf 50\% \\
    \midrule
    \multicolumn{4}{c}{\em WMT14 En$\Rightarrow$De} \\
    \midrule
    \textproc{Transformer} & \bf 27.8 & 19.1 & 11.5 \\
    \textproc{Lm4Mt} & \bf 27.8 & \bf 20.2 & \bf 13.6 \\
    \midrule
    \multicolumn{4}{c}{\em WMT14 De$\Rightarrow$En} \\
    \midrule
    \textproc{Transformer} & 31.3 & 19.9 & 11.8 \\
    \textproc{Lm4Mt} & \bf 31.5 & \bf 20.6 & \bf 14.2 \\
    \bottomrule
    \end{tabular}
    \vspace{-5pt}
\end{wraptable}

\paragraph{Model Robustness} 
To investigate the robustness of \textproc{Lm4Mt} to noisy inputs, we followed~\cite{Harshil:2018:GenerativeNMT,Zheng:2020:Mirror-Generative} to construct noisy test examples by omitting some words in the source-language sentences. Table~\ref{tab:miss-word} lists the results on WMT14 En$\Leftrightarrow$De test sets. It is evident that \textsc{Lm4Mt} is more robust than \textsc{Transformer} to missing word noise, and the performance improvement generally goes up with the increase of missing ratio.
This may be attributed to the auto-encoding loss $\mathcal{L}^{\mathrm{\textproc{AE}}}$, which induces source-side reconstruction supervision that can help \textproc{Lm4Mt} better ``denoise'' noisy inputs.

\subsection{Pivot-Based Translation}
\label{sec:pivot}

\textproc{Lm4Mt} uses a unified decoder to represent the source- and target-language sentences, both of which are learned by causal self-attention networks. Accordingly, the representation gap between the source- and target-language sentences are much smaller in \textproc{Lm4Mt} than that in encoder-decoder models. We believe that the shared representation can help better transfer the knowledge between source- and target-language texts. 
We verify the research hypothesis in the pivot-based translation scenario~\cite{Kim:2019:Pivot}, where the pivot language severs as the intermediate output to transfer the knowledge from the source language to the target.

Formally, we have parallel corpora $\mathcal{B}(X, Y) = \left \{ \langle \bm{\mathrm{x}}_n, \bm{\mathrm{y}}_n \rangle \right \}_{n=1}^{N}$ and $\mathcal{B}(Y, Z) = \left \{ \langle \bm{\mathrm{y}}_m, \bm{\mathrm{z}}_m \rangle \right \}_{m=1}^{M}$, and we aim to translate sentences of language $X$ into language $Z$. Assume that parallel corpus $\mathcal{B}(X, Z)$ is not available. To achieve this goal, we train NMT models on the mixture of $\mathcal{B}(X, Y)$ and $\mathcal{B}(Y, Z)$. Following \cite{johnson-etal-2017-googles}, we pend the target-language tag to the source sentence to indicate the translation direction for \textsc{Transformer}. For \textproc{Lm4Mt}, we use the tagging scheme illustrated in Figure~\ref{fig:lm4mt-train}. We also train \textsc{transformer} models using the same tagging scheme as \textsc{Lm4Mt} and find that only using the target-language tag works better for \textsc{Transformer}. At the inference time, we firstly translate the test data in language $X$ into the pivot language $Y$, which is then translated to language $Z$.
Intuitively, pivot-based translation tasks can benefit from \textsc{Lm4Mt} model that learns a shared representation across different languages.

\begin{table}[t]
    \setlength{\tabcolsep}{5.5pt}
    \caption{Pivot-based (English as the pivot language) translation performance measured by BLEU score. Results are reported on WMT14 test sets for En$\Leftrightarrow$De and En$\Leftrightarrow$Fr, and WMT20 test sets for De$\Leftrightarrow$Fr. NMT model for each direction is trained on the mxiture of WMT14 En-De and WMT14 En-Fr training corpora. Improvements over the \textsc{Transformer} baseline are highlighted in {\color{red!70} red cells} while deteriorations are represented in {\color{zptu!70} green cells}. Deeper color indicates larger performance gap.}
    \label{tab:pivot}
    \centering
    \begin{tabular}{r m{0pt} rrr m{0pt} rrr}
    \toprule
    \multirow{2}{*}{\bf Model} && \multicolumn{3}{c}{\bf De$\Rightarrow$En$\Rightarrow$Fr} && \multicolumn{3}{c}{\bf Fr$\Rightarrow$En$\Rightarrow$De} \\
    \cmidrule(lr){3-5}  \cmidrule(lr){7-9}
    && \bf De$\Rightarrow$En & \bf En$\Rightarrow$Fr & \bf De$\Rightarrow$Fr 
    && \bf Fr$\Rightarrow$En & \bf En$\Rightarrow$De & \bf Fr$\Rightarrow$De \\
    \cmidrule(lr){1-1}  \cmidrule(lr){3-5}  \cmidrule(lr){7-9}
    \textproc{Transformer-Base} && 30.8 & 39.5 & 25.7 && 35.2 & 26.8 & 21.4 \\
    \textproc{Lm4Mt-Base}       && \cellcolor{red!5} 30.9 & 39.5 & \cellcolor{red!45} \bf 26.6 && \cellcolor{red!45} 36.1 & \cellcolor{zptu!15} 26.5 & \cellcolor{red!70} \bf 22.8 \\
    \midrule
    \textproc{Transformer-Big} && 32.2 & 42.3 & 26.5 && 37.7 & 29.1 & 24.2 \\
    \textproc{Lm4Mt-Big}       && \cellcolor{red!15} 32.5 & \cellcolor{zptu!20} 41.9 & \cellcolor{red!60} \bf 27.7 && \cellcolor{red!20} 38.1 & \cellcolor{zptu!5} 29.0 & \cellcolor{red!45} \bf 25.1 \\
    \bottomrule
    \end{tabular}
\end{table}

In our experiments, we use English as the pivot language and aim to translate between De and Fr. We mix the WMT14 En-De and En-Fr corpora and the En-De corpus is upsampled to the same size with that of the En-Fr corpus. We use news-test2020 De$\Leftrightarrow$Fr to evaluate the translation performance. In each direction, all the models are trained only for the involved two intermediate directions. For instance, De$\Rightarrow$Fr models are trained on the mixture of De$\Rightarrow$En and En$\Rightarrow$Fr data.
Table~\ref{tab:pivot} shows the results of both base and big models. For pivot-based translation (i.e., De$\Leftrightarrow$Fr), \textproc{Lm4Mt} consistently outperforms the \textsc{Transformer} baseline by a large margin. To dispel the doubt that the superior performance of \textproc{Lm4Mt} might come from improvements accumulated in the two individual directions, we compute the performance gap between the two models in each direction, which is highlighted by color. The improvements in pivot-based translation directions are consistently larger than the two-step accumulated improvements, reconfirming the strength of \textproc{Lm4Mt} to exploit the shared knowledge between the source and target sides of parallel training data.

\subsection{Zero-Shot Translation}
\label{sec:zero-shot}

Previous studies have shown that a single multilingual NMT model can enable zero-shot translation -- translating between language pairs on which the NMT model has never been trained~\cite{johnson-etal-2017-googles,gu-etal-2019-improved}.
Compared with pivot-based translation in Section~\ref{sec:pivot}, zero-shot translation requires no pivot language as the intermediate output, thus requires better model representations to implicitly bridge between zero-shot language pairs.
We followed~\cite{johnson-etal-2017-googles} to conduct experiments with the multilingual setting. Specifically, we mix the bidirectional WMT14 En$\Leftrightarrow$De and En$\Leftrightarrow$Fr corpora to train a single multilingual NMT model, which is used to translate the zero-shot language pair of the WMT20 De$\Leftrightarrow$Fr test sets. 
Similarly to pivot-based experiments, we upsampled En$\Leftrightarrow$De corpora to the same size with that of the En$\Leftrightarrow$Fr corpora.

\begin{table}[t]
    \caption{Translation performance in the multilingual setting. ``Zero-Shot" means translating between unseen language pairs while ``Multilingual" means translating between seen language pairs.}
    \centering
    \label{tab:zero-translation-in-multilingual}
    \begin{tabular}{r rrrr rr}
    \toprule
    \multirow{2}{*}{\bf Model} & \multicolumn{4}{c}{\bf Multilingual} & \multicolumn{2}{c}{\bf Zero-Shot} \\
    \cmidrule(lr){2-5}  \cmidrule(lr){6-7}
     &  \bf De$\Rightarrow$En   & \bf En$\Rightarrow$Fr & \bf Fr$\Rightarrow$En    & \bf En$\Rightarrow$De   &   \bf De$\Rightarrow$Fr & \bf Fr$\Rightarrow$De \\
    \cmidrule(lr){1-1}  \cmidrule(lr){2-5}  \cmidrule(lr){6-7}
    \textproc{Transformer-Base}      & 30.4 & 38.5 & 34.8 & 25.8 & 21.9 & 13.8 \\
    \textproc{\textproc{Lm4Mt-Base}} & 30.4 & 38.4 & 34.8 & 25.9 & \bf 24.9 & \bf 19.5 \\
    \cmidrule(lr){1-1}  \cmidrule(lr){2-5}  \cmidrule(lr){6-7}
    \textproc{Transformer-Big}      & 33.1 & 41.9 & 38.0 & 29.2 & 23.8 & 16.7 \\
    \textproc{\textproc{Lm4Mt-Big}} & 33.9 & 41.1 & 37.8 & 28.6 & \bf 29.5 & \bf 24.9 \\
    \bottomrule
    \end{tabular}
\end{table}

\begin{table}[hbt!]
    \vspace{-15pt}
    \caption{Analyses of off-target translation issue.}
    \label{tab:off-target-language}
    \subfloat[Translation-language accuracy for zero-shot.]{
    \label{tab:off-target-ratio}
    \centering
    \begin{tabular}{r rr}
    \toprule
    \multirow{2}{*}{\bf Model} & \multicolumn{2}{c}{\bf Zero-Shot}\\
    \cmidrule(lr){2-3}
    & \bf De$\Rightarrow$Fr & \bf Fr$\Rightarrow$De \\
    \cmidrule(lr){1-1} \cmidrule(lr){2-3}
    \textproc{Transformer-Base} & 88.3\% & 79.7\% \\
    \textproc{Lm4Mt-Base}  & \bf 97.9\% & \bf 97.6\% \\
    \cmidrule(lr){1-1} \cmidrule(lr){2-3}
    \textproc{Transformer-Big} & 87.8\% & 76.5\% \\
    \textproc{\textproc{Lm4Mt-Big}} & \bf 98.8\% & \bf 98.5\% \\
    \bottomrule
    \end{tabular}}
    \hfill
    \subfloat[BLEU on sentences with the correct language.]{
    \label{tab:right-lang}
    \setlength{\tabcolsep}{3pt}
    \centering
    \begin{tabular}{r rr}
    \toprule
    \multirow{2}{*}{\bf Model} & \multicolumn{2}{c}{\bf Zero-Shot}\\
    \cmidrule(lr){2-3}
    & \bf De$\Rightarrow$Fr & \bf Fr$\Rightarrow$De \\
    \cmidrule(lr){1-1} \cmidrule(lr){2-3}
    \textproc{Transformer-Base} & 24.4 & 18.1 \\
    \textproc{Lm4Mt-Base}  & \bf 25.5 & \bf 19.9 \\
    \cmidrule(lr){1-1} \cmidrule(lr){2-3}
    \textproc{Transformer-Big} & 26.1 & 21.8 \\
    \textproc{\textproc{Lm4Mt-Big}} & \bf 29.4 & \bf 25.7 \\
    \bottomrule
    \end{tabular}}
\end{table}

Table~\ref{tab:zero-translation-in-multilingual} lists the results.
In zero-shot translation directions, \textsc{Transformer} performs much worse than \textsc{Lm4Mt}.
These results demonstrate the superiority of \textsc{Lm4Mt} on zero-shot translation. By manually checking the generated translations, we found that the failed models suffer from the {\em off-target translation issue} (i.e., translating into a wrong target language), which is the major source of the inferior zero-shot performance~\cite{zhang-etal-2020-improving}.
We follow \cite{zhang-etal-2020-improving} to employ the {\tt langdetect} library\footnote{\url{https://github.com/Mimino666/langdetect}} to detect the language of model outputs, and measure the translation-language accuracy for zero-shot cases.
Table~\ref{tab:off-target-ratio} lists the results, which provide empirical support for our findings. This is potentially caused by that \textproc{Transformer} tends to memorize translation directions seen in the training data while \textproc{Lm4Mt} can better generalize into unseen directions. More detailed results on off-target mistakes can be found in Appendix (Tables~\ref*{tab:detailed-off-target-ratio} and~\ref*{tab:detailed-off-target-case}).

To rule out the effect of the target language, we also evaluate the model performance on test examples that can be translated into the correct language by both \textsc{Transformer} and \textsc{Lm4Mt}. The results are shown in Table~\ref{tab:right-lang}, which demonstrate that \textsc{Lm4Mt} performs better than \textsc{Transformer} even when both of them can output the correct language in zero-shot cases.
We attribute the improvement to the reason that \textsc{Lm4Mt} can map sentences that come from various languages into a more unified representation space, thus can better understand the input sentences.

To further investigate why \textsc{Lm4Mt} shows a strong performance in zero-shot translation, we train \textsc{Lm4Mt} using different loss functions.
Table~\ref{tab:zero-shot-ablation} shows the results, indicating that the source-language supervision $\mathcal{L}^{\mathrm{\textproc{AE}}}$ is crucial for \textsc{Lm4Mt} to achieve good performance.
Training with $\mathcal{L}^{\mathrm{\textproc{AE}}}$, \textsc{Lm4Mt} is optimized by gradients induced from both the source- and target-side tokens, while encoder-decoder NMT models are trained only with the target-side loss, which may increase the representation gap between the source and target texts. Moreover, $\mathcal{L}^{\mathrm{\textproc{AE}}}$ enables better understanding of source sentences, which may prevent \textsc{Lm4Mt} from capturing spurious correlations between input sentences and language tags, which has been found to be harmful to zero-shot translation~\cite{gu-etal-2019-improved}.

\begin{table}[t]
    \caption{Effect of auto-encoding loss on zero-shot translation.}
    \centering
    \setlength{\tabcolsep}{4.8pt}
    \label{tab:zero-shot-ablation}
    \begin{tabular}{r r rrrr rr}
    \toprule
    \multirow{2}{*}{\bf Model} & \multirow{2}{*}{\bf Loss} & \multicolumn{4}{c}{\bf Multilingual} & \multicolumn{2}{c}{\bf Zero-Shot} \\
    \cmidrule(lr){3-6}  \cmidrule(lr){7-8}
    & &  \bf De$\Rightarrow$En   & \bf En$\Rightarrow$Fr & \bf Fr$\Rightarrow$En    & \bf En$\Rightarrow$De   &   \bf De$\Rightarrow$Fr & \bf Fr$\Rightarrow$De \\
    \midrule
    \textproc{Trans.-Base} & $\mathcal{L}^{\mathrm{\textproc{MT}}}$ & 30.4 & 38.5 & 34.8 & 25.8 & 21.9 & 13.8 \\
    \hdashline
    \multirow{3}{*}{\textproc{\textproc{Lm4Mt-Base}}} & $\mathcal{L}^{\mathrm{\textproc{MT}}}$ & 29.5 & 37.9 & 34.2 & 25.4 & 7.9 & 8.6 \\
    & $\mathcal{L}^{\mathrm{\textproc{AE}}}+\mathcal{L}^{\mathrm{\textproc{MT}}}$ & 29.3 & 37.5 & 33.7 & 24.7 & 23.8 & 18.6 \\
    & $\lambda_d\mathcal{L}^{\mathrm{\textproc{AE}}}+\mathcal{L}^{\mathrm{\textproc{MT}}}$ & 30.4 & 38.4 & 34.8 & 25.9 & \bf 24.9 & \bf 19.5 \\
    \midrule
    \textproc{Trans.-Big} & $\mathcal{L}^{\mathrm{\textproc{MT}}}$ & 33.1 & 41.9 & 38.0 & 29.2 & 23.8 & 16.7 \\
    \hdashline
    \multirow{3}{*}{\textproc{\textproc{Lm4Mt-Big}}} & $\mathcal{L}^{\mathrm{\textproc{MT}}}$ & 33.5 & 40.9 & 37.0 & 28.6 & 19.3 & 21.8 \\
    & $\mathcal{L}^{\mathrm{\textproc{AE}}}+\mathcal{L}^{\mathrm{\textproc{MT}}}$ & 33.0 & 40.5 & 37.1 & 27.6 & 28.8 & 23.3 \\
    & $\lambda_d\mathcal{L}^{\mathrm{\textproc{AE}}}+\mathcal{L}^{\mathrm{\textproc{MT}}}$ & 33.9 & 41.1 & 37.8 & 28.6 & \bf 29.5 & \bf 24.9 \\
    \bottomrule
    \end{tabular}
\end{table}

\section{Conclusion}

We propose a novel \textproc{Lm4Mt} model to perform machine translation using only a single language model. Although \textproc{Lm4Mt} is more simplified than the standard encoder-decoder NMT model, it can achieve competitive results with several strong encoder-decoder NMT baselines on standard machine translation tasks. 
\textproc{Lm4Mt} shows superior performance in both pivot-based and zero-shot translation scenarios by better transferring knowledge across languages with a unified representation.
One potential limitation of this work is that we only conduct experiments on parallel data. We believe that using additional monolingual data can further augment the performance of \textproc{Lm4Mt}, which can naturally consume monolingual texts without data augmentation methods. Another interesting direction is to investigate the effect of \textproc{Lm4Mt} on unsupervised NMT, which uses unified representations to bridge between language pairs without training signals from parallel data.

\bibliography{ref.bib}
\bibliographystyle{alpha}

\clearpage

\appendix

\section{Appendix}

\begin{table}[hbt!]
    \caption{Target-language ratios for zero-shot cases. The expected language is highlighted in {\color{zptu} green}.}
    \vspace{5pt}
    \label{tab:detailed-off-target-ratio}
    \centering
    \begin{tabular}{r rrrr rr}
    \toprule
    \multirow{2}{*}{\bf Model} & \multicolumn{3}{c}{\bf De$\Rightarrow$Fr} & \multicolumn{3}{c}{\bf Fr$\Rightarrow$De} \\
    \cmidrule(lr){2-4}  \cmidrule(lr){5-7}
     &  \bf De   & \bf En & \color{zptu} \bf Fr   & \bf Fr   &   \bf En & \color{zptu} \bf De \\
    \cmidrule(lr){1-1}  \cmidrule(lr){2-4}  \cmidrule(lr){5-7}
    \small \textproc{Transformer-Base} & 0.6\% & 9.8\% & \color{zptu} 88.3\% & 4.0\% & 15.2\% & \color{zptu} 79.7\% \\
    \small \textproc{Lm4Mt-Base} & 0.3\% & 1.4\% & \color{zptu} \bf 97.9\% & 0.9\% & 0.1\% & \color{zptu} \bf 97.6\% \\
    \cmidrule(lr){1-1} \cmidrule(lr){2-7}
    \small \textproc{Transformer-Big} & 2.8\% & 8.6\% & \color{zptu} 87.8\% & 12.7\% & 9.3\% & \color{zptu} 76.5\% \\
    \small \textproc{\textproc{Lm4Mt-Big}} & 0.1\% & 0.6\% & \color{zptu} \bf 98.8\% & 0.2\% & 1.0\% & \color{zptu} \bf 98.5\% \\
    \bottomrule
    \end{tabular}
\end{table}

\begin{table}[h!]
    \caption{Example for zero-shot De$\Rightarrow$Fr translation in multilingual translation. The \textsc{Transformer} model often translates into other language (e.g. {\color{rred} English} rather than French or {\color{bblue} copying} the source sentence), while our \textsc{Lm4Mt} suffers less from these mistakes.}
    \label{tab:detailed-off-target-case}
    \centering
    \begin{tabular}{r m{11cm}}
    \toprule
    \bf Source & Im Süden und Osten Europas tun sich die ökologischen Parteien nach wie vor schwer.  \\
    \hdashline
    \bf Reference & Les partis écologiques ont du mal à percer dans le sud et dans l'est de l’Europe. \\
    \midrule
    \bf \textproc{Trans.-Base} & \color{rred} In the south and east of Europe , the environmental parties are still struggling. \\
    \hdashline
    \bf \textproc{Lm4Mt-Base} & Dans les pays de l'Europe du Sud et de l'Est , les partis écologiques restent difficiles. \\
    \midrule
    \bf \textproc{Trans.-Big} & \color{rred} In the south and east of Europe , the environmental parties are still struggling. \\
    \hdashline
    \bf \textproc{Lm4Mt-Big} & Dans les pays du sud et de l'est de l'Europe, les partis écologiques continuent de se heurter à des difficultés. \\
    \bottomrule
    \bf Source & Bundesverteidigungsministerin Ursula von der Leyen ( CDU ) soll neue Präsidentin der EU-Kommission werden  \\
    \hdashline
    \bf Reference & La ministre fédérale de la Défense Ursula von der Leyen ( CDU ) doit devenir la nouvelle présidente de la Commission europé. \\
    \midrule
    \bf \textproc{Trans.-Base} & {\color{bblue}Bundesverteidigungsministerin} Ursula von der Leyen ( CDU ) {\color{bblue}soll} zum neuen {\color{bblue}Präsidenten der EU-Kommission werden}. \\
    \hdashline
    \bf \textproc{Lm4Mt-Base} & La ministre fédérale de la défense , Ursula von der Leyen ( CDU ) , est censée devenir la nouvelle présidente de la Commission européenne. \\
    \midrule
    \bf \textproc{Trans.-Big} & {\color{bblue}Bundesverteidigungsministerin} Ursula von der Leyen ( CDU ) {\color{bblue}soll neue Präsidentin der EU-Komission werden}. \\
    \hdashline
    \bf \textproc{Lm4Mt-Big} & Le ministre fédéral de la défense Ursula von der Leyen ( CDU ) doit devenir la nouvelle présidente de la Commission européenne. \\
    \bottomrule
    \end{tabular}
\end{table}

\end{document}